\documentclass[runningheads]{llncs}
\usepackage{graphicx}
\usepackage{amsmath,amssymb} 
\usepackage{color}
\usepackage{subcaption}
\usepackage{hyperref}
\hypersetup{%
  colorlinks=true,
  pdfborderstyle={/S/U/W 1}
}
\begin{document}
\pagestyle{headings}
\mainmatter

\def\ACCV18SubNumber{538}  

\title{Fast Video Shot Transition Localization with Deep Structured Models} 

\titlerunning{Fast Video Shot Transition Localization with Deep Structured Models} 

\authorrunning{Shitao Tang, Litong Feng, Zhanghui Kuang, Yimin Chen, Wei Zhang} 

\author{Shitao Tang\textsuperscript{1}, Litong Feng\textsuperscript{1}, Zhanghui Kuang\textsuperscript{1}, Yimin Chen\textsuperscript{1}, Wei Zhang\textsuperscript{1}} 
\institute{\textsuperscript{1}Sensetime Research}

\maketitle

\begin{abstract}
Detection of video shot transition is a crucial pre-processing step in video analysis. Previous studies are restricted on detecting sudden content changes between frames through similarity measurement and multi-scale operations are widely utilized to deal with transitions of various lengths. However, localization of gradual transitions are still under-explored due to the high visual similarity between adjacent frames. Cut shot transitions are abrupt semantic breaks while gradual shot transitions contain low-level spatial-temporal patterns caused by video effects in addition to the gradual semantic breaks, e.g. dissolve. In order to address the problem, we propose a structured network which is able to detect these two shot transitions using targeted models separately. Considering speed performance trade-offs, we design the following framework. In the first stage, a light filtering module is utilized for collecting candidate transitions on multiple scales. Then, cut transitions and gradual transitions are selected from those candidates by separate detectors. To be more specific, the cut transition detector focus on measuring image similarity and the gradual transition detector is able to capture temporal pattern of consecutive frames, even locating the positions of gradual transitions. The light filtering module can rapidly exclude most of the video frames from further processing and maintain an almost perfect recall of both cut and gradual transitions. The targeted models in the second stage further process the candidates obtained in the first stage to achieve a high precision. With one TITAN GPU, the proposed method can achieve a 30\(\times\) real-time speed. Experiments on public TRECVID07 and RAI databases show that our method outperforms the state-of-the-art methods. In order to train a high-performance shot transition detector, we contribute a new database ClipShots, which contains 128636 cut transitions and 38120 gradual transitions from 4039 online videos. ClipShots intentionally collect short videos for more hard cases caused by hand-held camera vibrations, large object motions, and occlusion. The database is avaliable at \hyperlink{https://github.com/Tangshitao/ClipShots}{https://github.com/Tangshitao/ClipShots}.
\end{abstract}


\section{Introduction}

\begin{figure}

\begin{subfigure}[b]{1.0\textwidth}
\centering
\includegraphics[height=0.4in,width=0.8\textwidth]{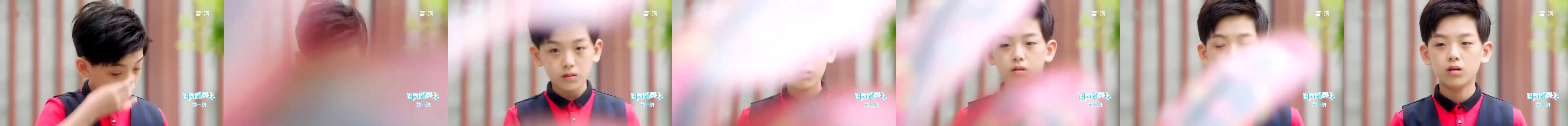}
\caption{Occlusion}
\end{subfigure}
\begin{subfigure}[b]{1.0\textwidth}
\centering
\includegraphics[height=0.4in, width=0.8\textwidth]{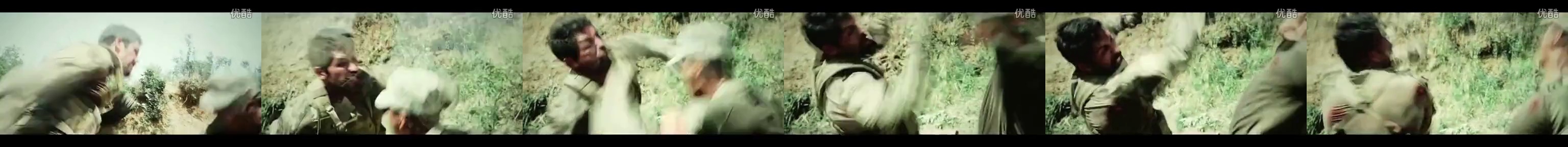}
\caption{Large motion}
\end{subfigure}
\begin{subfigure}[b]{1.0\textwidth}
\centering
\includegraphics[height=0.4in, width=0.8\textwidth]{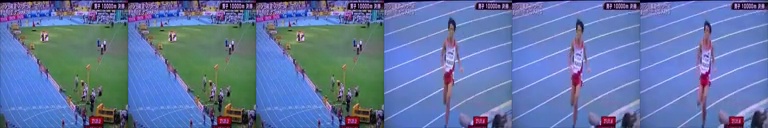}
\caption{Cut transition}
\end{subfigure}
\centering
\begin{subfigure}[b]{1.0\textwidth}
\centering
\includegraphics[height=0.4in, width=0.8\textwidth]{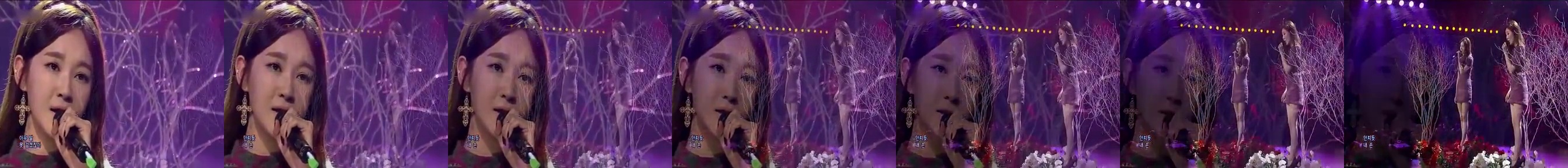}
\caption{Gradual transition}
\end{subfigure}

\caption{Challenge of shot boundary detection}\label{fig:1}
\end{figure}

Shot transition detector is a necessary component in many video recognition tasks. The goal of shot transition detection is to find semantic breaks in videos. Cut transitions are defined as abrupt transitions from one sequence to another while gradual transitions are almost the same but in a gradual manner. They share one common attribute, the start of a transition and the end of a transition are semantically different. Previous methods focus on finding both cut transitions and gradual transitions with one similarity function.\cite{yusoff2000video,yuan2005unified} Such methods have shown a great success in cut transition detection in the aspects of both speed and accuracy. However, when applied to gradual transition detection, it is not effective in the detection of gradual transitions. As Figure \ref{fig:1} shows, it is  widely recognized that many large motions or occlusion, e.g. camera movement, are detected as positive when only measuring similarity. In order to overcome this shortcoming, recent research\cite{lu2013fast,hassanien2017large} begins to explore the temporal pattern of gradual transitions. Therefore, in \cite{hassanien2017large}, the C3D ConvNet is adopted to classify segments into three classes (cut, gradual and background), which achieves state-of-the-art performance. Yet C3D ConvNet not only consumes too much computing resources, but is also not an effective architecture for handling both cut and gradual transitions, i.e. the lengths of gradual transitions are varying but C3D ConvNet is not designed for multi-scale detection. Inspired by this method and previous similarity measurement method, we present a cascade framework, consisting of a targeted cut transition detector and a targeted gradual transition detector. The cut transition detector, for measuring the image similarity, is fast and accurate while the gradual transition detector is capable of capturing the temporal pattern of gradual transitions in multi-scale level. In addition, compared to deepSBD, our framework can locate both cut transitions and gradual transitions accurately.

In this work, we present a new cascade framework, a fast and accurate approach for shot boundary detection. The first stage applies a ridiculously fast method to initially filter the whole video and selects the candidate segments. This stage is for accelerating the framework (up to 2 times faster than not) and facilitate the training for the cut/gradual detector. In the second stage, we use a well designed 2D ConvNet learning the similarity function between two images to locate the cut transitions. The third stage utilizes a novel C3D ConvNet model to locate positions of gradual transitions. Typically, we use the notation of default boxes introduced in \cite{liu2016ssd} and propose a novel single shot boundary detector (SSBD).

In sum, our framework is fast and accurate for shot boundary detection and achieves state-of-the-art performance on many public databases running at 700FPS without any bells and whistles.

Current datasets, i.e. TRECVID and RAI, are
not sufficient for training deep neural net due to limited dataset size. Besides, the training set is various in different work when evaluating supervised methods on TRECVID and RAI databases. For training a high performance neural network and a fair comparison between different methods, we contribute a new large-scale video shot database ClipShots consisting of different types of videos collected from Youtube and Weibo. ClipShots is the first large-scale database for shot boundary detection and will be released.

Aspects of novelty of our work include: 
\begin{itemize}
\item We separate cut transition detection and gradual transition detection, designing targeted network structures with different purposes. 
\item We design a cascade framework for accelerating the processing speed. 
\item We collect the first large-scale database for shot boundary detection training and evaluation.
\end{itemize}

\section{RELATED WORK}
In this section, we introduce the work related to our proposed framework.

\textbf{Unsupervised shot boundary detection method} In decades, many researchers explore to design similarity function finding transitions with hand-crafted features. In \cite{yusoff2000video}, average Intensity Measurement(AIM), Histogram Comparison(HC), Likelihood Ratio(LR) is used as the feature extractor. It is observed that similarities often vary gradually within a shot but abruptly in shot boundaries so the paper proposes an adaptive threshold should be applied when selecting positive samples. This method greatly improves the gradual transition performance compared to methods that only use static thresholds. Besides, another benefit is that it runs very fast so we integrate it in our framework to select potential shot boundaries. Yuan et al.\cite{yuan2005unified} proposes a graph partition model to perform temporal data segmentation. It treats every frame as a node and calculate the similarity metrix and the scores of the cuts, selecting feasible cuts whose scores are the local minima of the corresponding neighborhoods. These two methods all rely on well designed hand-crafted features to calculate the similarity of two images. 

\textbf{Supervised shot boundary detection method} Due to the shortcoming of unsupervised methods, Yuan et al.\cite{yuan2007formal} adopts a supervised way, a support vector machine trained to classify different shot boundaries with extracted features. In \cite{liu2007t}, shot boundaries are classified into 6 categories (cut, fast dissolve, fade in, fade out, dissolve, wipe). Different features are used to train different SVMs targeting at different shot boundaries. Researchers explore which features can most effectively classify the shot boundaries.

\textbf{Shot boundary detection with deep learning} Hassanien et al.\cite{hassanien2017large} introduces a simple C3D network that takes a segments of fixed length as input and classify it into 3 categories (cut, gradual, background). This method shows the effectiveness of ConvNet in this task. However, this method deals with gradual transitions of different scales in the same way and cannot locate the accurate`boundaries. Gygli\cite{gygli2017ridiculously} also adopts fully convolutional network. It takes the whole video sequence as input and assigns the positive label to the frames in transitions.

\textbf{Image similarity comparison} Image similarity computation is a necessary component in shot boundary detection. Deep learning has been successful on image similarity comparison task. In \cite{zagoruyko2015learning}, three architectures are proposed to compute image similarities, siamese net, image concatenation net, pseudo-siamese net. Empirical experiments show the image concatenation network and its variants obtain the best performance. In \cite{wang2014learning}, a ranking model that employs deep learning techniques to learn similarity metric directly from images. We apply the similarity measurement only for the cut transition detection.

\textbf{Object detection} State-of-the-art methods for general object detection are mainly based on deep ConvNet to extract rich semantic features from images. Liu et al.\cite{liu2016ssd} introduces single shot detector(SSD) using default boxes to match the feature to ground truth and achieve the speed of 19-46 fps. Our gradual detection model design share the same spirit with SSD.

\textbf{Action recognition} Recently, researchers have paid more attention on video recognition and  temporal detection. Carreira and Zisserman\cite{kay2017kinetics} has released the kinectics database for large-scale action classification. I3D\cite{carreira2017quo} shows a good weights initialization is necessary to train the C3D network. Qiu et al.\cite{qiu2017learning} proposes a fast network architecture based a spatial convolution kernel and temporal kernel to explore the temporal information. Action recognition is closely related to our work because we want to use temporal information to distinguish large motions and the gradual transitions.

\textbf{Action detection} This task focuses on learning how to detect action instances in untrimmed videos where the boundaries and categories of action instances have been annotated. Recently, many approaches adopt 'detection by classification' framework. Xu et al.\cite{xu2017r} builds faster-RCNN style architecture for fast classifying and locating actions. It first selects potential segments with region proposal network and proposes the ROI 3D pooling layer to extract rich features for further classification. In \cite{lin2017single}, the single shot detector locates action on feature map extracted from well trained action classification ConvNets. Escorcia et al.\cite{escorcia2016daps} proposes to generate a set of proposals based on the RNN network. Zhao et al.\cite{zhao2017temporal} models the temporal structure of each action instance via a structured temporal pyramid. Although some of the methods can be applied to gradual detection directly, these methods rely on extracting rich spatial-temporal features from a heavy ConvNet body, so these methods are far slower than our proposed methods. 

\begin{figure}
\centering
\includegraphics[height=3.4in,width=0.8\textwidth]{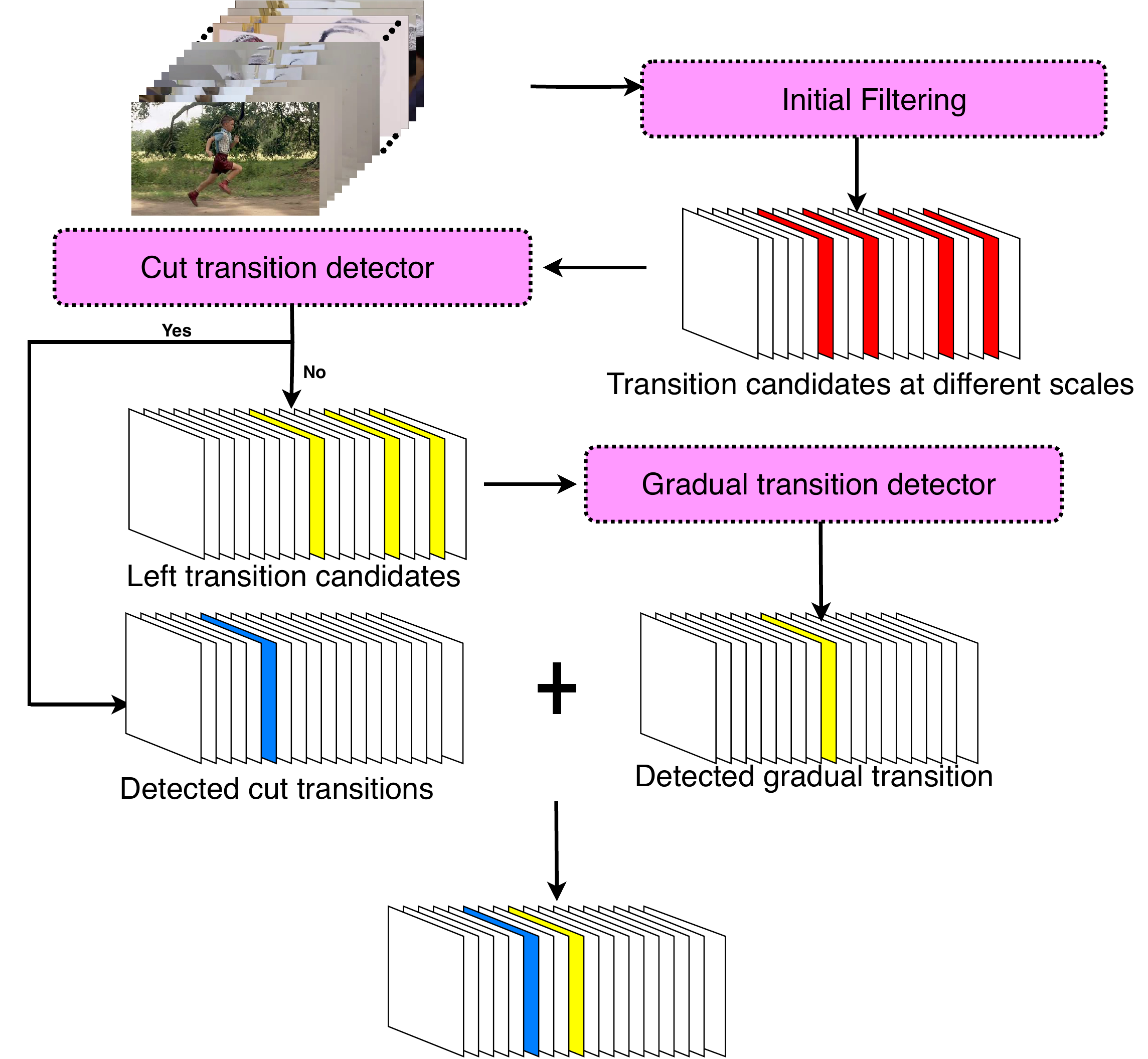}
\caption{An overview of our framework}\label{fig:2}
\end{figure}

\section{OUR APPROACH}
In this section, we will introduce our approach in details. The framework of our approach is shown in Figure \ref{fig:2}.

\subsection{An Overview}
The framework takes a video as input and predicts the locations of transitions. The proposed method, as shown in Figure \ref{fig:2}, is composed of three modules, including initial filtering, cut transition detector and gradual transition detector, implemented with three stages. 1) Adaptive thresholding produces a set of transition candidates. Each candidate comes with a center frame index indicating whether the content in frames has drastic changes. These positions may be transitions or caused by large motion, e.g. camera movement. 2) The candidate transitions are further feed into a strong cut transition detector to filter out false cut transitions. 3) For the remaining center frames which have negative responses to the cut detector, we expand them by \(x\) frames on both forward and backward temporal directions to form candidate segments. The gradual transition detector processes all these segments, locating the gradual transitions. The whole framework is designed in a cascade way and computation is light except the detection of gradual transition in the stage three. However, considering that most of the candidates have been filtered by the cut model, the number of candidates left for the gradual model is quite small and the computation at this stage is subtle.

\subsection{Initial Filtering}
As most of the consecutive video frames are highly similar to each other, a trivial unsupervised algorithm can be applied to reduce the candidate regions for further processing. A fast method, adaptive thresholding, is chosen as the initial filtering step. 

Let \(I_n\) and \(I_{n+1}\) be the potential transition candidates and \(F_{n-a+1}\), \(F_{n-a+2}\), ..., \(F_{n+a}\) be a set of features extracted from consecutive video frames in a sliding window of length \(2a\) centered at frame \(n\). In practice, we use the feature extracted from SqueezeNet\cite{iandola2016squeezenet} trained on Imagenet\cite{deng2009imagenet}. The computation cost in this step is subtle. We calculate the similarity metric of each frame \(S_i\), which is represented as the cosine distance between the current frame feature and its neighboring frame feature. Given the similarity metric of these frames as \(S_{n-a+1}\), \(S_{n-a+2}\), \(S_{n-a+3}\), ..., \(S_{n+a-1}\), the threshold of a window is calculated as
\begin{equation}
    T=t+\frac{\sigma}{n}\sum_{i=n-a+1}^{n+a-1}(1-S_i)
\end{equation}
 The hyper-parameter \(\sigma\) is the dynamic threshold ratio and \(t\) is the static threshold. In practice, we set \(\sigma\) to 0.05 and \(t\) to 0.5. The frame is selected as a candidate center if \(1-S_n\) is larger than \(T\). Lengths of gradual transitions vary greatly. In order not to miss any gradual transition, we down-sample frames with multiple temporal scales. At scale \(\omega\), we sample one video frame every \(\omega\) frames and do the above thresholding operations on these down-sampled frames. Finally, results of different scales are merged together. If two candidates on different scales are too close, i.e., within a distance of 5 frames. The candidate with a lower scale will be kept. In practice, we use scales of 1, 2, 4, 8, 16, and 32.

\subsection{Cut Model}\label{sec:1}
 Some image pairs are semantically similar even when they are cut transitions, i.e. images containing the same object but the backgrounds are different. Therefore, a stronger cut transition detector is needed for filtering out these negative cut candidates from the candidates selected by adaptive threholding. Zagoruyko and Komodakis\cite{zagoruyko2015learning} show CNN can learn the similarity function directly from image pairs. We design a ConvNet to determine whether a image pair is a cut transition or not. In this paper, we compare four models, including siamese, image concatenation, feature concatenation and C3D ConvNet. In contrast to deepSBD, where the position of the cut transition is unknown in one segment, adaptive thresholding can find the cut transition position accurately since it selects the pair of adjacent frames with the largest dissimilarity as the center, facilitating the learning task for our cut detector.

\textbf{Siamese} A siamese neural network consists of twin networks that accept distinct images and output their features. The parameters are shared between the twin networks and each network computes the same function, so two extremely similar images could not be mapped to very different location in the feature space. An energy loss function is added to the top for optimization. Besides, the network is symmetric, so that whenever we present two distinct images to the twin networks, the top conjoining layer will compute the same metric as if we were to present the same two images but to the opposite twins. In our problem, we choose contrastive loss as the top energy function. The siamese net outputs a similarity score. At inference, we select the score above some threshold. 

\textbf{Feature concatenation} This network can be seen as a variant of siamese network. More specifically, it has the structure of the siamese net described above, computing the feature using the same network architecture and weights. The loss energy function is not applied directly to the features. Instead, we concatenate features from both images and add cross entropy loss function to the top.

\textbf{Image concatenation} We simply consider the two patches of an RGB image pairs as a 6-channel image and feed it to a generic network. This network provides greater flexibility compared to the above models as it starts by processing the two patches jointly. It is fast to train and infer. Further more, it allows to concatenate multiple images as a input. We find the performance is much improved when using more images.

\textbf{C3D ConvNet} Hassanien et al.\cite{hassanien2017large} shows the C3D ConvNet is capable of classifying cut transitions. Therefore, we also test this structure for comparison. However, the C3D ConvNet is more complex than 2D ConvNet, which requires much computation resources. 

\begin{figure}
\centering
\includegraphics[height=3.4in, width=0.8\textwidth]{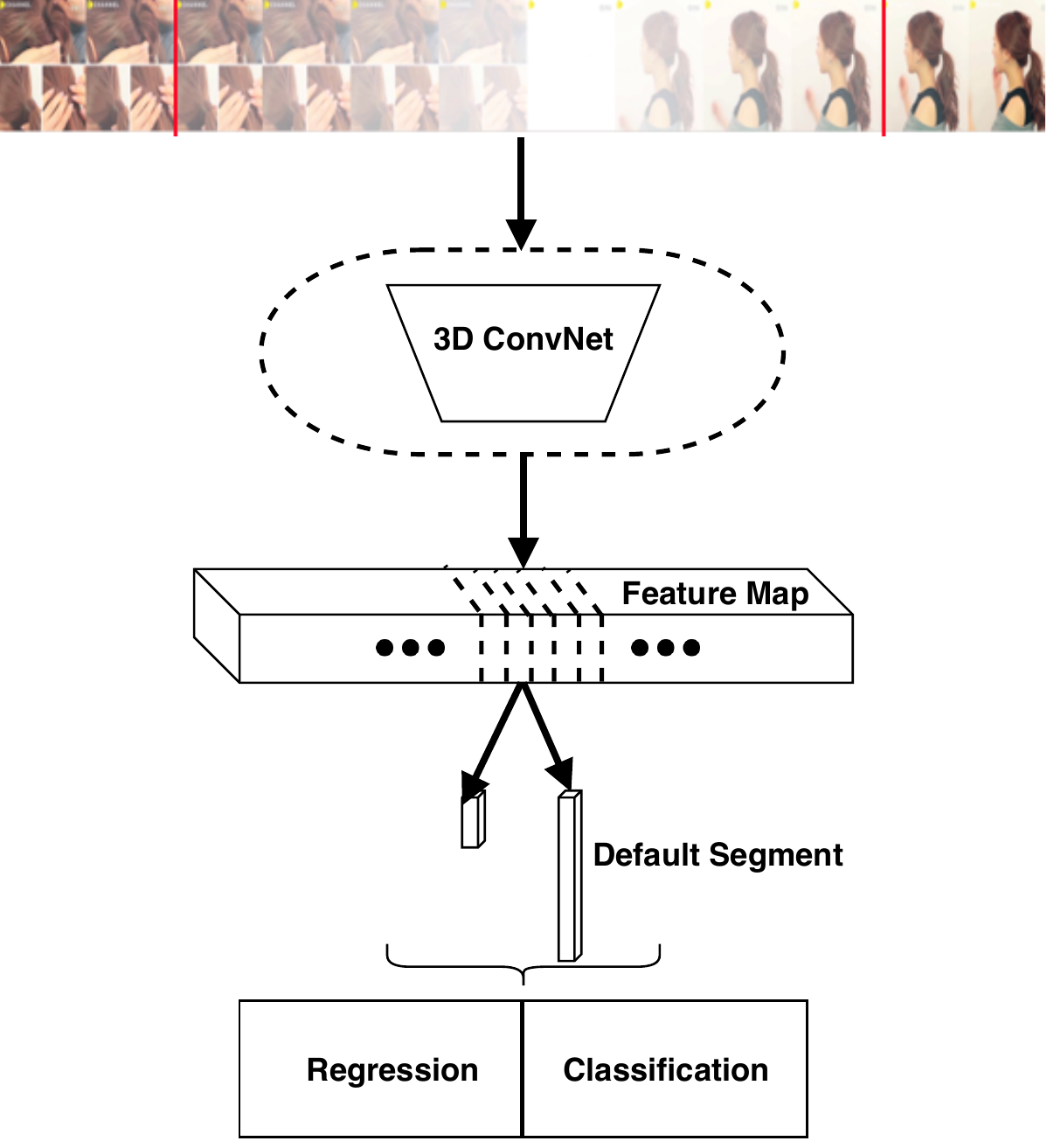}
\caption{An overview of gradual detector}\label{fig:3}
\end{figure}

\subsection{Gradual Model}
Inspired by region proposal network\cite{renNIPS15fasterrcnn} and single shot detector\cite{liu2016ssd}, we propose a single shot boundary network, a novel network to locate gradual transitions in a continuous video stream. The network, illustrated in Figure \ref{fig:3}, consists of 2 components, a shared C3D ConvNet feature extractor and subnets for classification and localization.

\textbf{Feature hierarchies} Innovated by deepSBD, the C3D ConvNet shows impressive performance in this task. Therefore, we use a C3D ConvNet to extract rich temporal feature hierarchies from a given input video buffer. The input to our model is a sequence of RGB video frames with a dimension of \(3\times L\times H\times W\) and we use ResNet-18 proposed in \cite{hara3dcnns} as the backbone network. However, unlike \cite{hara3dcnns}, the input to our model can be of variable lengths. We modify all the temporal strides to 1 in ResNet-18 so that the length of the final feature map is also L. The number of frames L can be arbitrary and is only limited by memory.

\textbf{Subnets for Classification and Location} Since the lengths of gradual transitions are various, we use the same notion default boxes introduced in \cite{liu2016ssd}. In our task, we call it default segments. Default segments are predefined multi-scale windows centered at a location. we put one default segment every \(l\times (1-a)\) frames where \(l\) is the length of the default segment and \(a\) is the positive IOU threshold. Therefore, each ground truth whose length is between \(l/a\) and \(l\times a\) can be matched to a default segment. The total number of default segments is \(L/(l\times(1-a))\). The default segments serve as reference segments for ground truth matching. To get features for predicting gradual transitions, we first apply a spatial global average pooling layer to reduce the spatial dimension to \(1\times1\).  At each location which has \(k\) default segments, we apply a \(2k\times3\times1\times1\) filter \(A\) for binary classification, and a \(2k\times3\times1\times1\) filter \(B\) for location refinement. For both \(A\) and \(B\), 3 is the size of the temporal convolution kernel. For \(A\), 2 corresponds to binary classification of a gradual transition or not. For B, 2 corresponds to two relative offsets of \(\{\delta c_i ,\delta l_i\}\) to the center location and the length of each default segment respectively, where the ground truth of \(\{\delta c_i ,\delta l_i\}\) is defined as
\begin{align}
\delta c_i&=(c-c_i)/l_i \\
\delta l_i&=log(l/l_i)
\end{align}
The mark \(c_i\) and \(l_i\) are the center location and the length of default segments while c and l is the ground truth position and length.

\textbf{Optimization strategy} In training, positive/negative labels are assigned to default segments. Following the same protocol in object detection, positive labels are assigned if default segments are overlapped with some ground truth if intersection of union \(IOU>a\) and negative labels are assigned for default segments if \(IOU<b\). Segments with IOU between \(a\) and \(b\) are ignored during training. In practice, we set \(a\) to 0.5  and \(b\) to 0.1, which achieves the best performance. As the length of the gradual transitions in our training data ranges in 3 to 40, we use 2 default segments of length 6 and 20 to cover all true transitions. Similar to single shot detector, we implement hard negative example mining and dynamically balance the positive and negative examples with a ratio of \(1:1\) during training. To utilize the GPU efficiently, we fixed the length of each segment, consisting of L consecutive frames, i.e., L is 64 in our experiment.

We train the network by optimizing the classification and the regression losses jointly with a fixed learning rate of 0.001 for 5 epochs. We adopt softmax loss for classification and smooth L1 loss for regression. The loss function is given in (\ref{Eq:locationloss}). The hyper-parameter \(\lambda\) is set to 1 in practice. \(Y_i^1\) is the predicted score and \(T_i^1\) is the assigned label. \(Y_i^2 =\{ \delta c_i ,\delta l_i\}\) is the predicted relative offset to the default segments and \(T_i^2\) is the target location. The loss function is the same as \cite{liu2016ssd}, which is
\begin{equation}\label{Eq:locationloss}
Loss=\frac{1}{N_{cls}} \sum_{i} L_{cls}(Y_i^1,T_i^1)+\lambda \frac{1}{N_{loc}} \sum_{i} L_{loc}(Y_i^2,T_i^2)
\end{equation}

\textbf{Inference} At inference, the framework processes input videos of varying lengths. However, in order not to exceed the limit of memory, a video will be divided into segments of length \(T_{seg}\) with a overlap of \(\frac{1}{2}T_{seg}\) such that transitions won't be missed due to the division. After predicting one video, we apply non maximum suppression(NMS) to all the predictions. If two predicted gradual transitions are overlapped, we remove the one with lower classification score.

\section{ClipShots}
Current datasets, i.e. TRECVID and RAI, are not sufficient for training deep neural network due to a limited size. In addition, previous work utilized different training sets when evaluating their supervised methods on TRECVID and RAI. Therefore, a benchmark is made for comparing different methods fairly. ClipShots is the first large-scale dataset for shot boundary detection collected from Youtube and Weibo covering more than 20 categories, including sports, TV shows, animals, etc. In contrast to TRECVID2007 and RAI, which only consist of documentaries or talk shows where the frames are relatively static, we construct a database containing 4039 short videos from Youtube and Weibo. Many short videos are home-made, with more challenges, e.g. hand-held vibrations and large occlusion. The training set consists of 3539 videos, 122760 cut transitions, and 35698 gradual transitions while the evaluation set consists of 500 videos, 5876 cut transitions, and 2422 gradual transitions. The types of these videos are various, including movie spotlights, competition highlights, family videos recorded by mobile phones etc. Each video has a length of 1-20 minutes. The gradual transitions in our database include dissolve, fade in fade out, and sliding in sliding out. In order to annotate such a large dataset, we design an annotation tool allowing annotators to watch multiple frames on a single page and select the begin frame and the end frame of transitions. More details are given in the appendix. Every video is double annotated for quality assurance.

\section{Experiments}
\subsection{Databases and Evaluation Metrics}
We will introduce the databases and evaluation metrics in this section.

\textbf{Training and evaluation set} The proposed framework is trained and tested on ClipShots. 
In order to illustrate the effectiveness of our approach and ClipShots, we also evaluated them on two public databases (TRECVID2007, RAI).

\textbf{Evaluation metrics} For all 3 databases, we use the standard TRECVID evaluation metrics: one-to-one match if the predicted boundary has at least 1 frame overlapped with the ground truth. For our testing set, we add an additional criterion using IOU to measure the localization performance. We assess performance quantitatively using precision (P), recall (R) and F-score (F). 

\subsection{Experiments Configuration}
We adopt adaptive thresholding to find candidate segments and adjust the parameters to make sure it achieves nearly 100\% recall for both cut and gradual transitions. For cut detector, 122760 positive examples and 224312 negative examples are used for training. For gradual detector, the training set contains 35698 ground truths. The potential segments filtered by adaptive thresholding are divided into subsegments of fixed length 64, with overlapped length of 32 between 2 consecutive segments. We choose ResNet-18 3D- ConvNet as the backbone, setting all the strides in the temporal dimension to 1 so that the temporal length of the output feature is identical with the input length. The weights of 3D ResNet-18 are initialized with model pretrained on kinectics database, as the inflated 3D-Conv \cite{carreira2017quo}. For both cut and gradual model, the positive examples and negative examples are highly unbalanced so the positive and negative samples are dynamically balanced with ratio 1:1 in each mini-batch.

\subsection{Experiments on ClipShots}

\begin{table}
\centering
\begin{tabular} {c|c|c|c}
\hline Model&P&R&F\\
\hline Image concat(2 frames)&0.771&0.793&0.782\\ 
\hline Image concat(4 frames)&0.775&0.862&0.816\\ 
\hline Image concat(6 frames)&0.776&0.934&0.848\\ 
\hline Feature concat(2 frames)&0.231&0.574&0.329\\ 
\hline Siamese(2 frames)&0.638&0.852&0.729\\ 
\hline C3D(16 frames)&0.760&0.910&0.831\\
\hline
\end{tabular}
\caption[Comparison of cut models]{Comparison of cut models. Image concat(6 frames) obtains the best performance.}\label{table:1}
\end{table}

\textbf{Cut detector comparison} In this section, we choose four potential models introduced in Sec. \ref{sec:1} and test their performance. We use ResNet-50 as the backbone for all models and a fixed learning rate of 0.0001, We train each model for 5 epochs. For C3D, we adopt the same configuration as deepSBD. For image concatenation model, we evaluated it with different number of images. We expand \(\emph{x}\) frames to the forward and backward in the temporal direction. As Table \ref{table:1} shows, the image concatenation model obtains best performance among these four models when using 4 or more frames. Siamese net performs worse than image concatenation (2 frames) and C3D network. Given the fact that siamese net cannot explore information on multiple frames and its computation cost is much larger than image concatenation, this architecture is not adopted in our framework. C3D network (16 frames) is a little better than image concatenation (2 frames), but much worse than image concatenation (4 frames or 6 frames). Feature concatenation is not a working architecture, but we still list it here. For image concatenation, we also study the relationship between the number of input images and performance. More input frames can improve performance. The model gains improvements when increasing the frame number from 2 to 6 and saturates around 6. Therefore, we use an input of 6 frames in our method considering both performance and the processing speed. 

\begin{table}
\centering
\begin{tabular} {c|c|c|c}
\hline method&initial filtering&cut&gradual\\
\hline (1)&No&DeepSBD&DeepSBD\\
\hline (2)&Yes&DeepSBD&DeepSBD\\
\hline (3)&Yes&image concat(6 frames)&DeepSBD\\
\hline (4)&Yes&image concat(6 frames)&SSBD\\
\hline 
\end{tabular}
\caption[general comparison]{All methods under a unified viewpoint. Different cut models and gradual models are compared.}\label{table:8}
\end{table}
\vspace{-5em}
\begin{table}
\centering
\small
\begin{tabular} {c|c|c|c|c|c|c}
\hline \multicolumn{1}{c|}{methods}&\multicolumn{3}{|c|}{cut}&\multicolumn{3}{|c}{gradual}\\
\hline &P&R&F&P&R&F\\
\hline (1)&0.765&0.910&0.831&0.770&0.622&0.688\\
\hline (2)&0.757&0.902&0.823&0.699&0.810&0.750\\
\hline (3)&0.776&0.934&0.848&0.711&0.830&0.766\\
\hline (4)&0.776&0.934&0.848&0.840&0.904&0.870\\
\hline 
\end{tabular}
\caption[general comparison]{Performance of different methods. Our method (4) obtains the best performance in both cut transition detection and gradual transitions detection.}\label{table:2}
\end{table}
\vspace{-2em}
\begin{table}
\centering
\begin{tabular} {c|c|c|c|c|c|c}
\hline \multicolumn{1}{c|}{methods}&\multicolumn{3}{|c|}{cut}&\multicolumn{3}{|c}{gradual}\\
\hline &P&R&F&P&R&F\\
\hline deepSBD (Original)&0.731&0.921&0.815&0.837&0.386&0.528\\
\hline deepSBD (ResNet-18)&0.765&0.910&0.831&0.770&0.622&0.688\\
\hline FCN\cite{gygli2017ridiculously}&0.410&0.093&0.151&0.393&0.053&0.093\\
\hline DSM (ours)&0.776&0.934&0.848&0.840&0.904&0.870\\
\hline 
\end{tabular}
\caption[general comparison]{Benchmark in ClipShots}\label{table:5}
\end{table}

\textbf{Ablation study} We conduct ablation study with different options. The detailed setting is shown in Table \ref{table:8}. The difference is mainly at cut models, gradual models, and whether initial filtering is used. We also implement deepSBD but the post processing technology introduced in \cite{hassanien2017large} is abandoned for a fair comparison. We adopt 3D ResNet-18 as the backbone for both deepSBD and our single shot boundary detector. 

\textit{Method (1)} The model classifies segments directly into 3 categories (cut, gradual, and background).

\textit{Method (2)} Compared to method (1), initial filtering is utilized to find candidate segments for deepSBD. As is shown, the performance of gradual transition is higher than the original deepSBD. It is implied that the initial filtering can also improve performance of deepSBD.

\textit{Method (3)} For gradual transitions, the deepSBD model only classifies the segments into 2 categories (gradual transition and background) so cut transitions are treated as negative samples. For cut detector, we use image concatenation model. Compared to method(2), the results show the single shot boundary detector is better than deepSBD by a large margin given that the cut detector are the same.

\textit{Method (4)} The results reveals that our single shot boundary detector is far better than deepSBD. We attribute the performance gain to the following reasons: 1) The receptive field of our model is much bigger than deepSBD, hence the detector can exploit more temporal information. 2) Our default segment design is effective for dealing with gradual transitions of multi scales.

\textbf{Benchmark in ClipShots}
We implement \cite{hassanien2017large,gygli2017ridiculously} and evaluate them in ClipShots. Table \ref{table:5} summaries performance of different methods. DeepSBD with 3D ResNet-18 is significantly better than the original network (3D Alexnet alike). It is also noted that the method introduced in \cite{gygli2017ridiculously} obtains worst performance on our dataset. Its network only consists of 3 convolutional layers. We suppose it is too shallow to fit large-scale data. 

\begin{table}
\centering
\begin{tabular} {c|c}
\hline model&speed(FPS)\\
\hline deepSBD&382\\
\hline adaptive thresholding+deepSBD&680\\
\hline ours&700\\
\hline
\end{tabular}
\caption[Comparison of cut models]{Comparison of speed}\label{table:speed}
\end{table}

\textbf{Speed Comparison}
In this section, we compare the speeds of different models as shown in Table \ref{table:speed}. The code is implemented using PyTorch and tested with one TITAN XP GPU. Our method is nearly 2 times faster than the original deepSBD on account of adaptive thresholding based initial filter.

\begin{table}
\centering
\begin{tabular} {c|c|c|c}
\hline IOU&P&R&F\\
\hline 0.1&0.813&0.865&0.839\\ 
\hline 0.5&0.726&0.772&0.748\\ 
\hline 0.75&0.599&0.637&0.618\\ 
\hline
\end{tabular}
\caption[localization]{Localization performance. We calculate the F1-score at different IOU threshold.}\label{table:3}
\end{table}
\vspace{-1em}
\textbf{Gradual Model Localization Performance} An accurate localization of gradual transitions is important in many video recognition task. Therefore, we also evaluate performance of the gradual transition localization using the proposed framework. F1 scores are measured at different IOU level \((0.1,0.5,0.75)\). A predicted gradual transition is considered as correct only if its \(IOU>a\), otherwise it's considered wrong. When the IOU is 0.75, we can still obtain a F1 score of 0.618, indicating the proposed gradual detector is able to accurately locate gradual transitions.  

\begin{table}
\centering
\begin{tabular} {c|c|c|c|c|c|c}
\hline \multicolumn{1}{c|}{methods}&\multicolumn{3}{|c|}{cut}&\multicolumn{3}{|c}{gradual}\\
\hline &P&R&F1&P&R&F1\\
\hline ATT\cite{liu2007t}&0.996&0.979&0.972&0.802&0.709&0.753\\
\hline THU11\cite{yuan2007formal}&0.982&0.968&0.975&0.733&0.718&0.725\\
\hline Marburg\cite{muhling2007university}&0.942&0.945&0.944&0.595&0.766&0.670\\
\hline NHT\cite{kawai2007shot}&0.975&0.816&0.945&0.768&0.578&0.66\\
\hline Priya\cite{domnic2014walsh}&0.972&0.976&0.974&0.869&0.719&0.78\\
\hline DeepSBD\cite{hassanien2017large}&0.978&0.968&0.973&0.826&0.731&0.776\\
\hline Ours&0.971&0.988&0.980&0.813&0.806&0.810\\
\hline Ours (correct label)&0.981&0.997&0.989&0.838&0.845&0.841\\
\hline 
\end{tabular}
\caption[Trecvid comparison]{Trecvid07 top performers.}\label{table:4}
\end{table}
\vspace{-3em}
\subsection{Experiments on TRECVID07}
TRECVID07 contains a total of 17 videos, including 2236 cut transitions and 225 gradual transitions. They are all color and black/white documentaries. The videos include cases such as global illumination variation, smoke, fire, and fast non-rigid motion. We take the ground truth from TRECVID07 SBD task. In addition, the experimental results of the proposed method over this database are compared to the top performers of TRECVID07 SBD task. We find some of the ground truths are wrong, so we correct these labels. Evaluation results using original labels and corrected labels are both reported. The cut and gradual models are trained with the same training setting described in Section 5.2.

In Table \ref{table:4}, we present a comparative evaluation of the shot boundary detection performance with existing state-of-the-art approaches in terms of F1-score and report the results using both the original ground truth and the corrected ground truth. We evaluate cut transitions and gradual transitions separately. Cut transitions are the most part of all transitions in a video so it plays a dominate role in the overall performance. For cut transitions, we improve the-state-of-art by 0.6\%, which is a huge improvement considering there is no much space for improvement. In fact, the errors concentrate in black/white videos due to the lack of similar ones in the training set. Further improvement can be achieved through adding more black/white videos into the training set. For gradual transitions, we achieve 2.9\%  improvement comparing to the state-of-the-art when using the original ground truth and 6.4\% improvement when using the corrected ground-truth.
\vspace{-2em}
\begin{table}
\centering
\begin{tabular} {c|c}
\hline &F1 score \\
\hline Apostolidis et al.\cite{apostolidis2014fast}&0.84\\
\hline Baraldi et al.\cite{baraldi2015shot}&0.84\\ 
\hline Song et al.\cite{song2016click}&0.68\\ 
\hline Michael et al.\cite{gygli2017ridiculously}&0.88\\
\hline Hassanien et al.\cite{hassanien2017large}&0.934\\
\hline Ours&0.935\\
\hline
\end{tabular}
\caption[RAI comparison]{RAI comparison}
\end{table}
\vspace{-3em}
\subsection{Experiments on RAI}
RAI database is a collection of ten randomly selected broadcasting videos from the Rai Scuola video archive 1, which are mainly documentaries and talk shows. This database includes 722 cut transitions and 263 gradual transitions. Shots have been manually annotated by a set of human experts. The proposed method achieves a competitive results compared to deepSBD. It is noted that DeepSBD adopts posting-processing technology, i.e. filtering the segments whose HSV similarity under a threshold, which is not used in our methods. We perform evaluations on TRECVID and RAI using the same models, weights, and hyper-parameters, which indicates the proposed framework are robust on different databases.

\section{Conclusion}
We propose a cascade shot transition detection framework and annotate the first large-scale shot boundary database. Adaptive threshoding is adopted to find candidate regions for acceleration. The cut and gradual transition detector are designed separately. The cut transition detector is for measuring similarity while the gradual transition detector is for capturing temporal patterns. Especially, the gradual detector is able to locate gradual transitions of multi-scales. We outperform state-of-the-art methods on both TRECVID and RAI databases. In addition, our framework is very fast, achieving a 30\(\times\) real-time speed.

\bibliographystyle{splncs}
\bibliography{citation}

\begin{thebibliography}{10}

\bibitem{yusoff2000video}
Yusoff, Y., Christmas, W.J., Kittler, J.:
\newblock Video shot cut detection using adaptive thresholding.
\newblock In: BMVC. (2000)  1--10

\bibitem{yuan2005unified}
Yuan, J., Li, J., Lin, F., Zhang, B.:
\newblock A unified shot boundary detection framework based on graph partition
  model.
\newblock In: Proceedings of the 13th annual ACM international conference on
  Multimedia, ACM (2005)  539--542

\bibitem{lu2013fast}
Lu, Z.M., Shi, Y.:
\newblock Fast video shot boundary detection based on svd and pattern matching.
\newblock IEEE Transactions on Image processing \textbf{22} (2013)  5136--5145

\bibitem{hassanien2017large}
Hassanien, A., Elgharib, M., Selim, A., Hefeeda, M., Matusik, W.:
\newblock Large-scale, fast and accurate shot boundary detection through
  spatio-temporal convolutional neural networks.
\newblock arXiv preprint arXiv:1705.03281 (2017)

\bibitem{liu2016ssd}
Liu, W., Anguelov, D., Erhan, D., Szegedy, C., Reed, S., Fu, C.Y., Berg, A.C.:
\newblock Ssd: Single shot multibox detector.
\newblock In: European conference on computer vision, Springer (2016)  21--37

\bibitem{yuan2007formal}
Yuan, J., Wang, H., Xiao, L., Zheng, W., Li, J., Lin, F., Zhang, B.:
\newblock A formal study of shot boundary detection.
\newblock IEEE transactions on circuits and systems for video technology
  \textbf{17} (2007)  168--186

\bibitem{liu2007t}
Liu, Z., Gibbon, D., Zavesky, E., Shahraray, B., Haffner, P.:
\newblock At\&t research at trecvid 2007.
\newblock In: Proc. TRECVID Workshop. (2007)  19--26

\bibitem{gygli2017ridiculously}
Gygli, M.:
\newblock Ridiculously fast shot boundary detection with fully convolutional
  neural networks.
\newblock arXiv preprint arXiv:1705.08214 (2017)

\bibitem{zagoruyko2015learning}
Zagoruyko, S., Komodakis, N.:
\newblock Learning to compare image patches via convolutional neural networks.
\newblock In: Computer Vision and Pattern Recognition (CVPR), 2015 IEEE
  Conference on, IEEE (2015)  4353--4361

\bibitem{wang2014learning}
Wang, J., Leung, T., Rosenberg, C., Wang, J., Philbin, J., Chen, B., Wu, Y.,
  et~al.:
\newblock Learning fine-grained image similarity with deep ranking.
\newblock arXiv preprint arXiv:1404.4661 (2014)

\bibitem{kay2017kinetics}
Kay, W., Carreira, J., Simonyan, K., Zhang, B., Hillier, C., Vijayanarasimhan,
  S., Viola, F., Green, T., Back, T., Natsev, P.,  et~al.:
\newblock The kinetics human action video dataset.
\newblock arXiv preprint arXiv:1705.06950 (2017)

\bibitem{carreira2017quo}
Carreira, J., Zisserman, A.:
\newblock Quo vadis, action recognition? a new model and the kinetics dataset.
\newblock In: 2017 IEEE Conference on Computer Vision and Pattern Recognition
  (CVPR), IEEE (2017)  4724--4733

\bibitem{qiu2017learning}
Qiu, Z., Yao, T., Mei, T.:
\newblock Learning spatio-temporal representation with pseudo-3d residual
  networks.
\newblock In: 2017 IEEE International Conference on Computer Vision (ICCV),
  IEEE (2017)  5534--5542

\bibitem{xu2017r}
Xu, H., Das, A., Saenko, K.:
\newblock R-c3d: Region convolutional 3d network for temporal activity
  detection.
\newblock In: The IEEE International Conference on Computer Vision (ICCV).
  Volume~6. (2017) ~8

\bibitem{lin2017single}
Lin, T., Zhao, X., Shou, Z.:
\newblock Single shot temporal action detection.
\newblock In: Proceedings of the 2017 ACM on Multimedia Conference, ACM (2017)
  988--996

\bibitem{escorcia2016daps}
Escorcia, V., Heilbron, F.C., Niebles, J.C., Ghanem, B.:
\newblock Daps: Deep action proposals for action understanding.
\newblock In: European Conference on Computer Vision, Springer (2016)  768--784

\bibitem{zhao2017temporal}
Zhao, Y., Xiong, Y., Wang, L., Wu, Z., Tang, X., Lin, D.:
\newblock Temporal action detection with structured segment networks.
\newblock In: The IEEE International Conference on Computer Vision (ICCV).
  Volume~8. (2017)

\bibitem{iandola2016squeezenet}
Iandola, F.N., Han, S., Moskewicz, M.W., Ashraf, K., Dally, W.J., Keutzer, K.:
\newblock Squeezenet: Alexnet-level accuracy with 50x fewer parameters and< 0.5
  mb model size.
\newblock arXiv preprint arXiv:1602.07360 (2016)

\bibitem{deng2009imagenet}
Deng, J., Dong, W., Socher, R., Li, L.J., Li, K., Fei-Fei, L.:
\newblock Imagenet: A large-scale hierarchical image database.
\newblock In: Computer Vision and Pattern Recognition, 2009. CVPR 2009. IEEE
  Conference on, IEEE (2009)  248--255

\bibitem{renNIPS15fasterrcnn}
Ren, S., He, K., Girshick, R., Sun, J.:
\newblock Faster {R-CNN}: Towards real-time object detection with region
  proposal networks.
\newblock In: Advances in Neural Information Processing Systems ({NIPS}).
  (2015)

\bibitem{hara3dcnns}
Hara, K., Kataoka, H., Satoh, Y.:
\newblock Can spatiotemporal 3d cnns retrace the history of 2d cnns and
  imagenet?
\newblock arXiv preprint \textbf{arXiv:1711.09577} (2017)

\bibitem{muhling2007university}
M{\"u}hling, M., Ewerth, R., Stadelmann, T., Z{\"o}fel, C., Shi, B.,
  Freisleben, B.:
\newblock University of marburg at trecvid 2007: Shot boundary detection and
  high level feature extraction.
\newblock In: TRECVID. (2007)

\bibitem{kawai2007shot}
Kawai, Y., Sumiyoshi, H., Yagi, N.:
\newblock Shot boundary detection at trecvid 2007.
\newblock In: TRECVID. (2007)

\bibitem{domnic2014walsh}
Domnic, S.:
\newblock Walsh--hadamard transform kernel-based feature vector for shot
  boundary detection.
\newblock IEEE Transactions on Image Processing \textbf{23} (2014)  5187--5197

\bibitem{apostolidis2014fast}
Apostolidis, E., Mezaris, V.:
\newblock Fast shot segmentation combining global and local visual descriptors.
\newblock In: Acoustics, Speech and Signal Processing (ICASSP), 2014 IEEE
  International Conference on, IEEE (2014)  6583--6587

\bibitem{baraldi2015shot}
Baraldi, L., Grana, C., Cucchiara, R.:
\newblock Shot and scene detection via hierarchical clustering for re-using
  broadcast video.
\newblock In: International Conference on Computer Analysis of Images and
  Patterns, Springer (2015)  801--811

\bibitem{song2016click}
Song, Y., Redi, M., Vallmitjana, J., Jaimes, A.:
\newblock To click or not to click: Automatic selection of beautiful thumbnails
  from videos.
\newblock In: Proceedings of the 25th ACM International on Conference on
  Information and Knowledge Management, ACM (2016)  659--668

\end{thebibliography}

\end{document}


\pagestyle{headings}
\mainmatter

\def\ACCV18SubNumber{***}  

\title{Appendix} 
\titlerunning{ACCV-18 submission ID \ACCV18SubNumber}
\authorrunning{ACCV-18 submission ID \ACCV18SubNumber}

\author{Anonymous ACCV 2018 submission}
\institute{Paper ID \ACCV18SubNumber}

\maketitle

\section{Annotation tool}
\begin{figure}
\centering
\includegraphics[height=3.4in,width=1.0\textwidth]{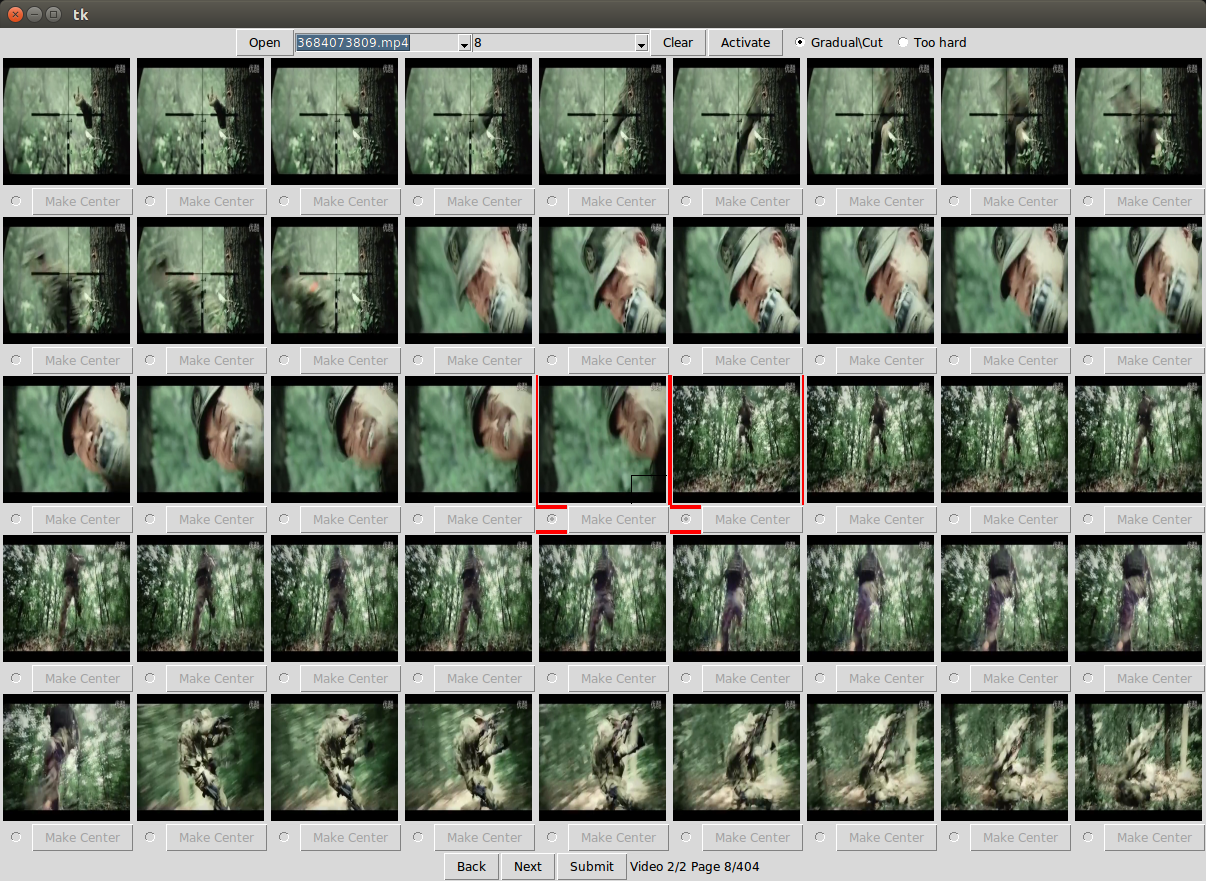}
\caption{Annotation tool interface}\label{fig:1}
\end{figure}

The Graphical User Interface (GUI) is shown in \ref{fig:1}. This tool allows the annotator to watch 45 consecutive frames of a video segment simultaneously. This visualization is helpful for annotators to find a transition with a single look. On each page, the annotator can annotate at most one transition, with the selection of a begin frame and a end frame. The selected begin frame and end frame are asked to cross the center frame on a page. This setting is for assuring an annotator can obtain enough frames to determine the ends of a transition. When a page covers several ground truths, the others can be made to the center frame by clicking the button 'Make Center'. In the top-right corner, there are two tags, 'Gradual/Cut' or 'Too hard'. 'Gradual/Cut' represents the user wants the selected segments to be the ground truth while 'Too hard' represents the segments is difficult to be determined. We exclude 'Too hard' segments in the training stage. The user can watch next or previous page by clicking the buttons in the bottom or selecting the page number in the top.